\documentclass[journal]{IEEEtran}

\usepackage{cite}

\ifCLASSINFOpdf
\else
\fi
\usepackage{graphicx}

\usepackage{amsmath}
\usepackage{amssymb}          

\usepackage{array}
\usepackage{placeins}

\usepackage{hyperref} 
\hypersetup{
    colorlinks=true,
    linkcolor=black,
    filecolor=black, 
    citecolor=blue,
    urlcolor=blue,
}

\begin{document}
%
\title{Programming Robot Behaviors with\\Execution Management Functions}
%
%
%

\author{Martin~Molina, Pablo~Santamaria and Abraham~Carrera
\thanks{Authors affiliation: Department of Artificial Intelligence, \textit{Universidad Politécnica de Madrid}, Spain. Research Group CVAR (Computer Vision and Aerial Robotics).}
}

\maketitle

\thispagestyle{empty}

\begin{abstract}
The control architecture of autonomous robots can be developed by programming and integrating multiple software components that individually control separate behaviors. This approach requires additional mechanisms to coordinate their concurrent execution. This paper presents a programming method for such components that has been designed to facilitate their coordinated execution. Each component is programmed as a module that controls a separate robot behavior together with a set of functions for execution management. The details of this proposal are formulated in the form of a ROS-based software library called \textit{behaviorlib}. This solution has been used to program general behavior controllers that have been successfully reused to build multiple applications in aerial robotics.

\end{abstract}


%

\section{Introduction}
%
%
%
%

\begin{figure}[htb!]
\begin{center}
\includegraphics[width=0.48\textwidth]{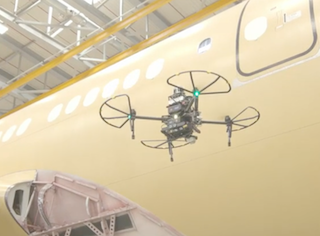}
\caption{Example of autonomous aerial robot in an inspection mission. The aerial robot navigates autonomously around an airplane and observes its surface with the help of a camera.}
\label{fig:dar system}
\end{center}
\end{figure}

\IEEEPARstart{A}{utonomous} robots exhibit multiple behaviors during the execution of missions. For example, Figure \ref{fig:dar system} shows an autonomous aerial robot\footnote{This robot was developed in our research group CVAR for the DAR project (\url{https://vimeo.com/393907228}).} in a visual inspection mission. In this type of mission, the aerial robot exhibits behaviors such as taking off, following a planned path, landing, recognizing objects visually, and avoiding dynamic obstacles.

The development of this type of robots can be done by programming software components that separately control each robot behavior using specialized algorithms (e.g., algorithms for motion control, motion planning, computer vision, etc.). This approach requires additional mechanisms to coordinate the concurrent execution of such components taking into account multiple aspects such as the robot's goals in a given mission, the current situation of the environment, and the relationships between concurrent behaviors (e.g., incompatibility or dependency). Such coordination mechanisms may be difficult to design and implement in complex autonomous robots and the selected solution for coordination significantly affects the quality of the final system in aspects such as performance efficiency and maintainability.

This paper presents a programming method for robot behavior controllers that has been designed to facilitate their coordinated execution. According to this method, each behavior is programmed as a module that includes specific functions for execution management (e.g., execution monitoring and activation management functions).

The details of this proposal are formulated in the form of a software library called \verb|behaviorlib| that is general to be used for programming behaviors of different types of robots. This library is implemented in C++ language using inter-processes mechanisms provided by ROS (Robot Operating System) and it is freely available in a public software repository\footnote{\url{https://github.com/cvar-upm/behaviorlib}}.

This solution has been successfully applied to build programs that implement general aerial robot behaviors. These programs are part of the Aerostack software environment \cite{Sanchez2017} and have been used in the development of multiple aerial robot applications.

The remainder of this paper is organized as follows. Section 2 describes work related to the proposal presented in this paper. Section 3 presents an overview of the approach described in this work. Section 4 describes the details of the ROS-based implementation. Section 5 shows how the method described in this paper has been used for building the set of reusable behavior controllers that are part of the Aerostack software framework for aerial robotics. Finally, Section 6 presents general conclusions about the present work.

\section{Related work}
\begin{figure*}[htb!]
\begin{center}
\includegraphics[width=0.85\textwidth]{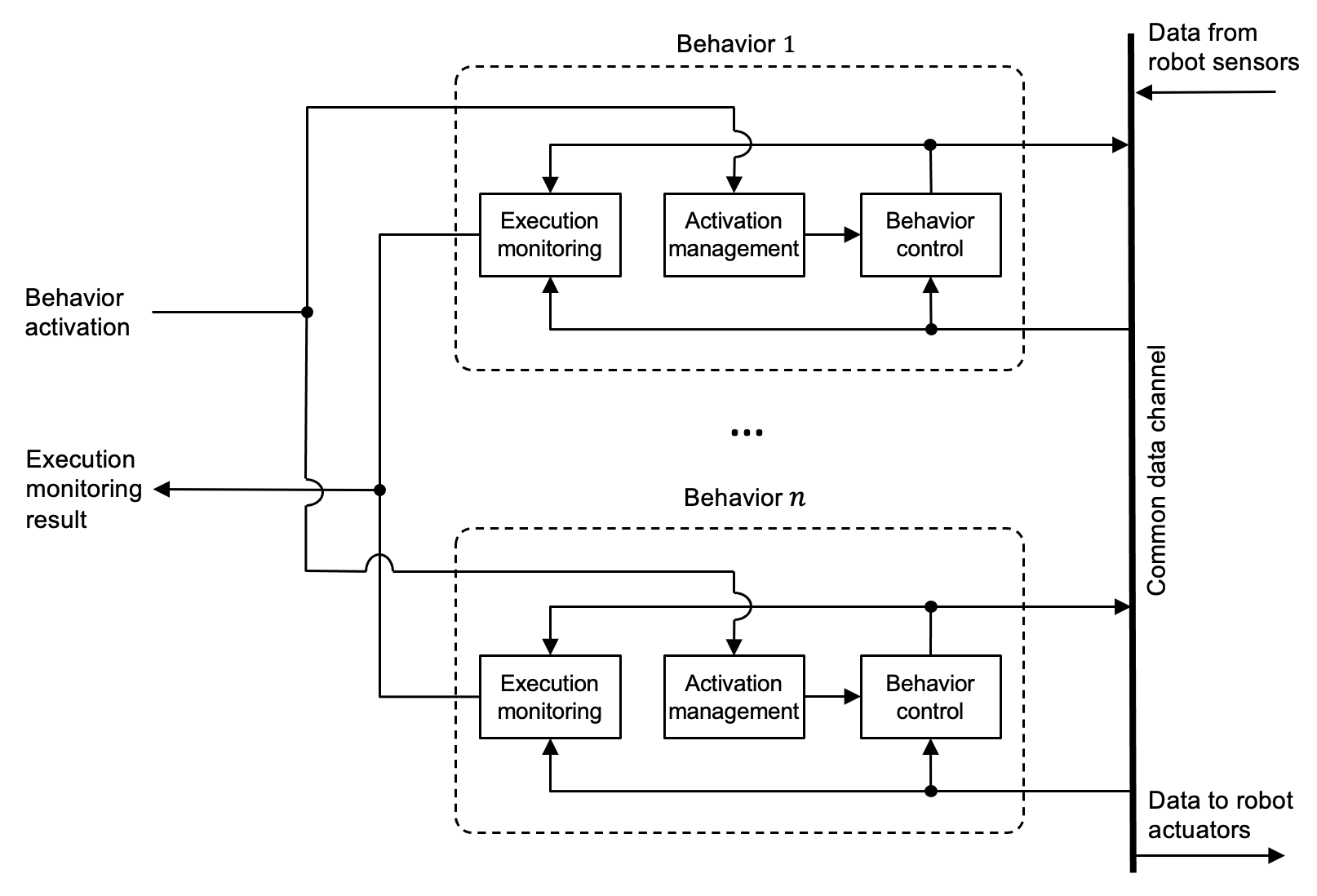}
\caption{Distributed organization of the functionalities associated to robot behaviors. Each behavior is implemented with three main functions: behavior control, activation management and execution monitoring.}
\label{fig:decentralized organization}
\end{center}
\end{figure*}

The idea of programming separate modules with uniform interfaces that are used as basic components of a control architecture is followed by multiple software tools in robotics. For example, the software toolbox CRAM \cite{Beetz2010} uses \textit{process modules} implemented as Lisp programs that encapsulate functions executing ROS nodes. The tool ROSPlan \cite{Cashmore2015} uses \textit{actions} that are implemented with the help of a C++ class (called \textit{RPActionInterfaceClass}). The software tool Genom \cite{Fleury1997} uses \textit{modules} that are generated automatically (in language C) for control architectures based on the LAAS architecture.  Another related software tool used in ROS is \verb|actionlib|. With this solution, a client communicates to a server the action to be done and the server communicates the result of the action and feedback reporting the incremental progress of that action.

In comparison with these tools, our method has been designed for programming a different type of component, a robot behavior controller (instead of an action, a general data processor or a state), which is used with a different type of interface and communication protocol.

A particular characteristic of our programming approach is that each robot behavior controller performs also behavior monitoring, which is an important function of autonomous systems \cite{Kephart2003}. For example, it helps to provide cognizant failure (e.g., fault detection) \cite{Gat1994} or the ability to explain the own actions \cite{Lomas2012}. 

Execution monitoring can be implemented in multiple ways (see for example \cite{Pettersson2005} for a review of methods of execution monitoring). Our solution proposes to distribute the supervision among the different behaviors of the robot. Jahn \textit{et al.} \cite{Jahn2019} also proposes the idea of a distributed approach for execution monitoring (besides other functions) using a \textit{reflective operator} \cite{Gausemeier2010}. In our work, we develop this idea in a different way (e.g., by identifying specific categories of executing monitoring functions) and we provide an open source software library to help programming these functions. 

\section{General approach for behavior programming}

The approach followed in this work for programming robot behaviors is influenced by the behavior-based paradigm in robotics\cite{Brooks1986} \cite{Arkin1998}. According to this paradigm, the global control is divided into a set of behavior controllers and each one is in charge of a specific control aspect separately from the other behavior controllers. A robot behavior\footnote{The programming method presented in this paper is designed for basic or primitive behaviors, i.e., simple behaviors that are used for building other more complex behaviors.} can be implemented using a controller together with a mechanism to activate and deactivate it. An example of a behavior controller is a robot motion controller that uses a control method, such as PID (proportional-integral-derivative) control or MPC (model predictive control). It operates in a loop receiving input data about the robot localization and generates output data to be used by motion actuators.

In our approach, we generalize the idea of behavior to manage basic robot skills related not only to physical actuators but also to data management (e.g., visual recognition, self localization, environment mapping, motion planning, etc.). In general, a behavior controller operates in a loop receiving input data from sensors (or data generated from other behavior controllers) and generating output data to be used by actuators (or to be used by other behavior controllers).

Regarding behavior execution goals, we distinguish between two types of robot behaviors\footnote{This division is also mentioned in the literature of robotics. For example, these two categories have been distinguished using the terms \textit{servo}, for recurrent behaviors, and \textit{ballistic} for goal-based behaviors \cite{Jones2002}.}: (1) \textit{goal-based} behaviors that are defined to reach a final state or attain a goal (for example, in an aerial robot, the behavior taking off), and (2) \textit{recurrent} behaviors that perform an activity recurrently or maintain a desired state (for example, a behavior for visual object recognition).

\subsection{Common data channel}

To organize the robot control architecture with multiple behaviors that operate concurrently, we assume that there is a common data channel that is shared among all behavior controllers (see Figure \ref{fig:decentralized organization}). To facilitate the interoperability of behavior controllers, the meaning of the data types used in the data channel should follow conventions defined for the whole control architecture (e.g., through the use of  standards for message types).

Each behavior controller should be able to execute separately assuming that the required input data is available in the common data channel. This is related to the the property called \textit{situated} behaviors, meaning that each behavior controller is designed to operate correctly only under certain assumed situations. The separate execution of behaviors facilitates the addition of new behaviors with flexibility, without affecting the implementation of other behaviors and global control mechanisms.

\subsection{Activation management}

Each robot behavior is programmed with an activation management mechanism. This mechanism handles how to start and stop the execution of the behavior controller (optionally, it could also handle other execution control mechanisms such as pause and resume execution). Activation management is useful, for example, to cope with incompatible behaviors or to manage 
efficiently robot resources. In an mobile robot, for instance, a robot motion controller following a planned path should be deactivated if an incompatible type of motion control needs to be activated (e.g., following a mobile object).

There are different software tools in robotics that use similar activation mechanisms to manage executable components. For example, ROS 2 uses \textit{managed nodes}, i.e., nodes with a managed life cycle to start and stop the execution of ROS nodes. FlexBE \cite{Schillinger2016} uses a similar idea applied to states.

\subsection{Execution monitoring}

Besides the two functions mentioned above (behavior control and activation management), we propose in this work that each behavior program includes an execution monitoring function (see Figure \ref{fig:decentralized organization}). 

Execution monitoring is a kind of self-awareness computing process \cite{Lewis2016} by which the robot observes and judges its own behavior. In the case of a program for a specific robot behavior, execution monitoring gathers and processes data about such behavior. The gathered data is used to assess dynamically the execution of the behavior controller by contrasting this data with expectations. This process is useful, for example, to assess the robot operation in relation to a dynamic environment to detect the presence of unexpected situations (e.g., failures).

In our approach, the knowledge used for execution monitoring is distributed among different behaviors. This is appropriate because execution monitoring normally uses behavior-specific knowledge. For example, the knowledge used to monitor the execution of a path following behavior is different from the knowledge used to monitor the execution of visual object recognition behavior.

More specifically, execution monitoring checks, for example, whether the current situation of the environment satisfies the assumptions for the behavior controller to operate correctly and how this situation may affect the behavior controller performance. This is related to the to the property of situated behaviors mentioned above. For example, in the case of a behavior controller that performs visual recognition, execution monitoring can check that the lighting assumptions in the environment are verified. In addition, it can estimate how the current lighting conditions affect the performance of visual recognition.

Execution monitoring also detects when the behavior has reached a prefixed goal. For example, in the case of a motion controller to follow a certain path, the monitoring function detects that the behavior has reached the goal when the robot arrives at the destination point. The analysis of the robot behavior in relation to the robot goals is also used for fault detection. For example, in the previous example, a failure can be detected if the robot does not arrive to the destination point in a maximum expected time or if the robot moves in the opposite direction to the direction that is expected to follow towards the destination point (considering absence of obstacles).

\begin{table}
\caption{Possible termination causes of behavior execution.}
\label{tab:termination_causes} 
\centering
\footnotesize
\begin{tabular}{ | m{1.1cm} | m{6.5cm} | } 
\hline
\textbf{Cause} & \textbf{Description} \\
\hline
Goal achieved &  The behavior execution has finished because the goal established by the task has been achieved. For example, if the behavior task is following a particular path, the goal is achieved when the robot reaches the final point of the path.  \\ 
\hline
Time out & The behavior execution has finished because it was not able to complete the task in the maximum expected time. \\ 
\hline
Wrong progress & The behavior execution has finished because the robot is not working as expected. This could happen, for example, if a robot moves in the opposite direction to the direction that is expected to be followed towards the destination point (considering absence of obstacles). \\ 
\hline
Situation change & The behavior execution has finished because the required conditions in the environment are not satisfied. For example, a behavior for visual recognition can finish in this way if the conditions about the illumination conditions are not satisfied. \\ 
\hline
Process failure & The behavior execution has finished because at least one of the software processes that execute the behavior is not working correctly. For example, this happens if a processes is blocked due to a programming bug that creates an infinite loop. \\ 
\hline
Interrupted & The behavior execution has finished because its deactivation has been requested.\\ 
\hline
\end{tabular}
\end{table}

The result of execution monitoring is generated locally for each behavior. In order to facilitate the coordination of multiple behaviors, this result can be expressed in a uniform way using a prefixed type of message. Table \ref{tab:termination_causes} shows a set of possible termination causes of behavior execution that may be considered according to the type of behavior implementation described in this paper.

\section{ROS-based library for behavior programming}

This section describes the ROS-based library \verb|behaviorlib| that was created to help programming robot behaviors following the approach described above. 

\begin{figure}[htb!]
\begin{center}
\includegraphics[width=0.50\textwidth]{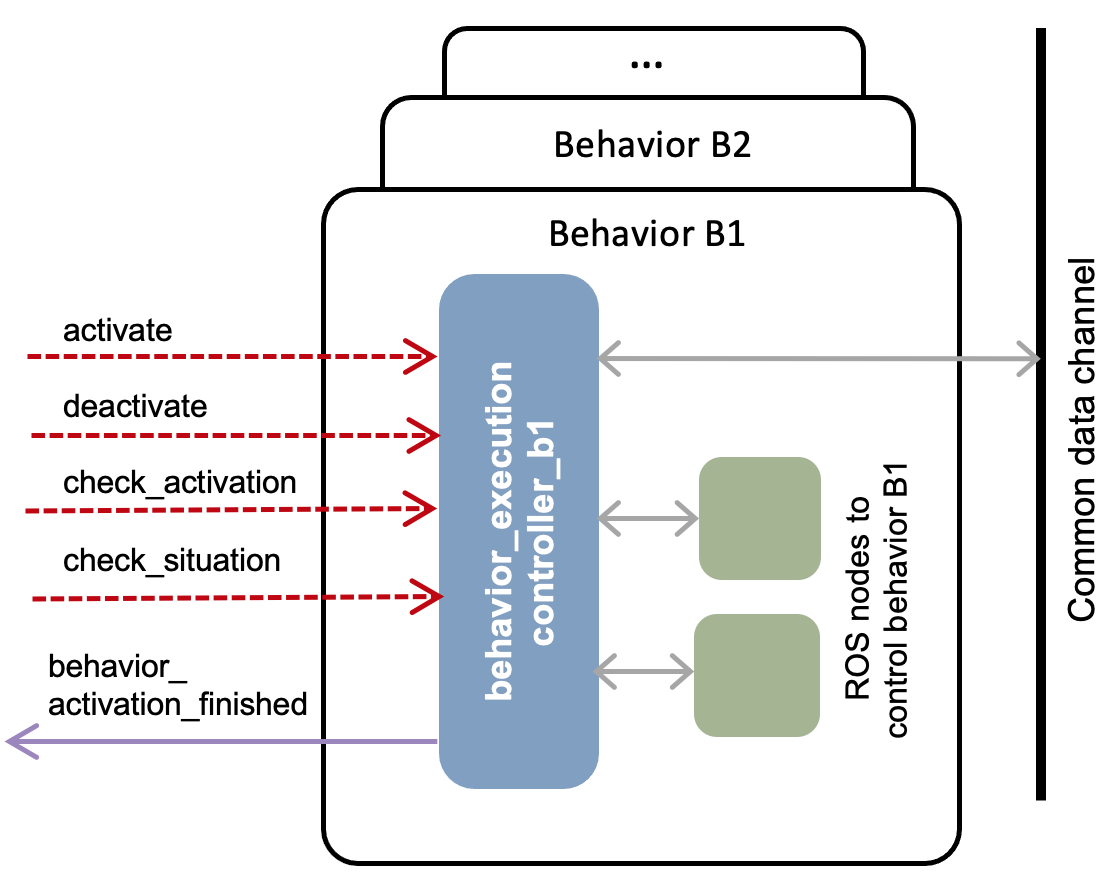}
\caption{ROS-based implementation of a robot behavior. In this figure, colored round rectangles are ROS nodes, dashed red arrows represent ROS services and the continuous purple arrow represents a ROS topic.}
\label{fig:ROS node}
\end{center}
\end{figure}

The basic component of this library is a type of ROS node called \textit{behavior execution manager} (see figure \ref{fig:ROS node}). For each robot behavior, there is one node of this type.

This node provides a uniform interface for the behavior to (1) activate and deactivate the behavior, (2) notify when the behavior finishes its execution and (3) inform about the termination cause. This interface is implemented using inter-process communication mechanisms used in ROS. The node provides the following request-reply ROS services:

\begin{itemize}
\item \verb|activate|: This service activates the execution of the behavior. It admits parameter values. For example, the behavior {\footnotesize\verb|ROTATE|} can be activated with a parameter value \verb|angle = 90| expressing that the robot should rotate 90 degrees.
\item \verb|deactivate|: This service terminates the behavior execution.
\item \verb|check_activation|: This service checks whether the behavior is active or not. If the behavior is active and was activated using some parameter values, this service also returns these parameter values.
\item \verb|check_situation|: This service verifies that the behavior can be activated in the current situation of the environment (for example, to activate the behavior take off, an aerial robot must be landed). This function can also estimate a value of behavior performance considering the current state of the environment.
\end{itemize}

The behavior execution manager publishes a message in the ROS topic called \verb|behavior_activation_finished|. This message is sent when the execution of the behavior has finished and contains the behavior identification and the cause of termination.

The behavior execution manager can also facilitate the communication with the common data channel used in the control architecture of the robotic system. This is useful, for example, when the behavior is implemented by reusing existing programs (e.g., general programs and software tools for robotics that are freely available as part of ROS). These programs can also use additional ROS nodes (represented as green nodes in Figure \ref{fig:ROS node}). In this case, the behavior execution manager operates as a \textit{wrapper} that converts input/output data used by the reused programs to the standard format used by the common data channel. 

\begin{figure}[htb!]
\begin{center}
\includegraphics[width=0.35\textwidth]{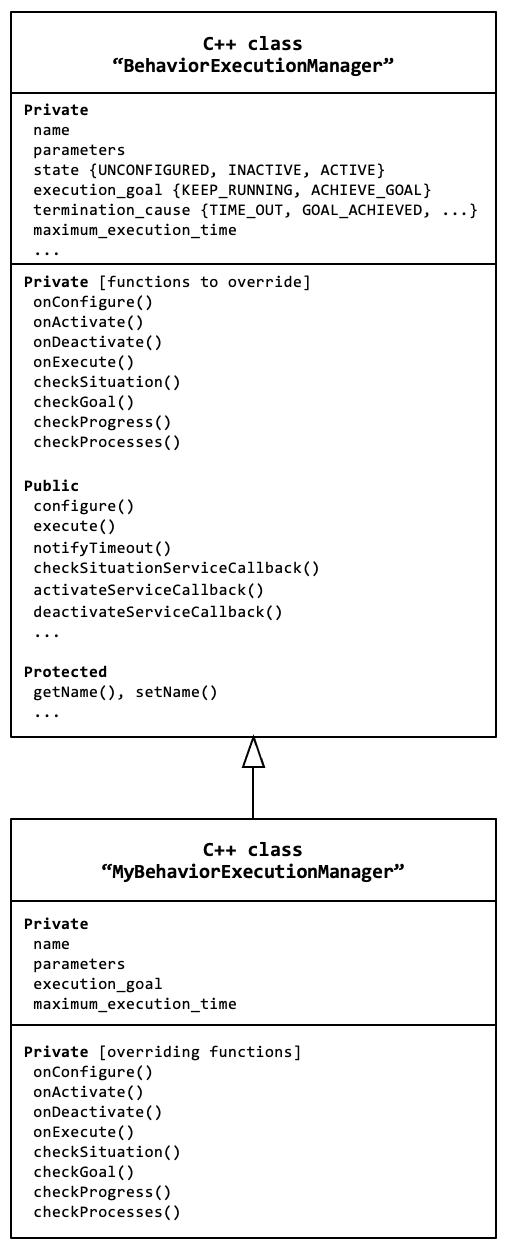}
\caption{C++ classes to program a behavior execution manager. A subclass is created for each particular robot behavior (for example the subclass {\scriptsize \textsf{MyBehaviorExecutionManager}}. Each subclass includes behavior-specific functions for execution monitoring and activation management that override functions defined in the general class.}
\label{fig:behavior execution controller class}
\end{center}
\end{figure}

To program each behavior execution manager, the library \verb|behaviorlib| provides a C++ class called \verb|BehaviorExecutionManager| (see Figure \ref{fig:behavior execution controller class}) together with types of ROS messages and services. A C++ subclass is created for each particular robot behavior (for example the subclass \verb|MyBehaviorExecutionManager|). Each subclass includes functions for the behavior that override some of the functions defined in the general class. These functions implement the specific procedures for each behavior related to execution monitoring (see Table \ref{tab:monitoring functions}) and activation management (see Table \ref{tab:activation control functions}). Each subclasss also includes values for certain attributes to help execution monitoring. For example, a particular value for the attribute \verb|maximum_execution_time| can be specified to establish an adequate timeout threshold for the particular behavior.

The class \verb|BehaviorExecutionManager| includes general functions that are shared by each subclass. For example the function \verb|execute()| calls the functions of Table \ref{tab:monitoring functions} to perform execution monitoring. This function is executed periodically using a frequency that can be adjusted for each particular behavior.

\begin{table}
\caption{Specific functions for behavior execution monitoring.}
\label{tab:monitoring functions} 
\centering
\footnotesize
\begin{tabular}{ | m{2.5cm} | m{5.1cm} | } 
\hline
\textbf{Function} & \textbf{Description} \\
\hline
\verb|checkSituation()| &  This function checks whether the current state of the environment satisfies the situation assumptions of the behavior. This function can also estimate dynamically a value of behavior performance considering the current situation of the environment (e.g., a real number  in the interval [0,1] where the maximum value 1 corresponds to the optimal conditions of the environment). \\ 
\hline
\verb|checkGoal()| &  This function checks whether the behavior has reached the goal. For example, this function detects that a behavior moving the robot to a certain destination has reached its goal when the robot arrives at its destination. \\ 
\hline
\verb|checkProgress()| &  This function checks whether the robot state is progressing correctly according to what is expected by the active behavior.\\ 
\hline
\verb|checkProcesses()| &  This function checks that the computational processes used to control the behavior are running as expected.  \\ 
\hline
\end{tabular}
\end{table}

\begin{table}
\caption{Specific functions for behavior activation management.}
\label{tab:activation control functions} 
\centering
\footnotesize
\begin{tabular}{ | m{2.3cm} | m{5.3cm} | } 
\hline
\textbf{Function} & \textbf{Description} \\
\hline
\verb|onConfigure()| &  This function reads configuration parameters (e.g., from files or from ROS parameters).  \\ 
\hline
\verb|onActivate()| & This function initiates inter-process communication (e.g., subscribe and advertise), ensures that processes are running, sets initial values for variables, publishes initial messages and calls initial services.  \\ 
\hline
\verb|onExecute()| & This function executes the next step of the iteration process.  \\ 
\hline
\verb|onDeactivate()| & Shutdown of inter-process communication ensuring a safe stable state. \\ 
\hline
\end{tabular}
\end{table}

\section{Application and evaluation} 
\begin{table}[hbt!]
\caption{Behavior systems used in Aerostack.}
\label{tab:behaviors Aerostack} 
\centering
\scriptsize
\begin{tabular}{ | m{1.9cm} | m{5.35cm} | } 
\hline 
\textbf{Behavior system} & \textbf{Behavior} \\
\hline & \verb|LAND| \\
\cline {2-2} & \verb|TAKE_OFF| \\
\cline {2-2} \verb|basic_|& \verb|WAIT| \\
\cline {2-2} \verb|quadrotor_| & \verb|SELF_LOCALIZE_WITH_GROUND_TRUTH| \\
\cline {2-2} \verb|behaviors| & \verb|ESTIMATE_POSITION_WITH_SENSOR| \\
\cline {2-2} & \verb|ESTIMATE_POSITION_WITH_LINEAR_SPEED| \\
\cline {2-2} & \verb|HOVER_WITH_FLIGHT_ACTION_CONTROL| \\
\hline \verb|multi_| & \verb|SELF_LOCALIZE_WITH_|\\
\verb|sensor_fusion| & \verb|EKF_SENSOR_FUSION| \\
\hline  \verb|quadrotor_| & \verb|FOLLOW_PATH_WITH_MPC_CONTROL| \\
\cline {2-2} \verb|motion_| & \verb|ROTATE_WITH_MPC_CONTROL| \\
\cline {2-2} \verb|with_mpc_| & \verb|HOVER_WITH_MPC_CONTROL| \\
\cline {2-2} \verb|control| & \verb|QUADROTOR_MPC_MOTION_CONTROL| \\
\hline  & \verb|FOLLOW_PATH| \\
\cline {2-2} \verb|quadrotor_|& \verb|ROTATE_WITH_PID_CONTROL| \\
\cline {2-2} \verb|motion_| & \verb|KEEP_HOVERING_WITH_PID_CONTROL| \\
\cline {2-2} \verb|with_pid_| & \verb|KEEP_MOVING_WITH_PID_CONTROL| \\
\cline {2-2} \verb|control| & \verb|QUADROTOR_PID_MOTION_CONTROL| \\
\cline {2-2} & \verb|QUADROTOR_PID_THRUST_CONTROL| \\
\hline \verb|quadrotor_| & \verb|MOVE_VERTICAL_WITH_| \\
\verb|motion_| & \verb|PLATFORM_CONTROL| \\
\cline {2-2} \verb|with_platform_| & \verb|MOVE_AT_SPEED_WITH_PLATFORM_CONTROL| \\
\cline {2-2} \verb|control| & \verb|ROTATE_WITH_PLATFORM_CONTROL| \\
\hline \verb|navigation_| & \verb|GENERATE_PATH_WITH_OCCUPANCY_GRID| \\
\cline {2-2} \verb|with_lidar| & \verb|CLEAR_OCCUPANCY_GRID| \\
\cline {2-2} & \verb|SAVE_OCCUPANCY_GRID| \\
\hline \verb|attention_to_| & \verb|PAY_ATTENTION_TO_| \\
\verb|visual_markers| & \verb|QR_CODES| \\
\hline  \verb|swarm_| & \verb|PAY_ATTENTION_TO_ROBOT_MESSAGES| \\
\cline {2-2} \verb|interaction| & \verb|INFORM_ROBOTS| \\
\cline {2-2} & \verb|INFORM_POSITION_TO_ROBOTS| \\
\hline \verb|operator_| & \verb|INFORM_OPERATOR| \\
\cline {2-2} \verb|interaction| & \verb|REQUEST_OPERATOR_ASSISTANCE| \\
\hline
\end{tabular}
\end{table}

This section presents examples that illustrate how the approach described in this paper has been successfully used for building applications in aerial robotics. This section also presents the results of experiments conducted
to evaluate how the programming method  affects the performance efficiency (e.g., processing time, CPU usage and memory consumption).

\subsection{Applications in aerial robotics}

The approach presented in this paper for behavior programming has been successfully used to develop a library of general behaviors for aerial robotics. This library has been published as part of the of the software environment Aerostack\footnote{\url{http://www.aerostack.org}} (Version 4.0, Auster distribution). Here, we describe how the method described in this paper was used for building the Aerostack's library of behaviors, showing practical guidelines that were applied to organize the different types of behaviors.

Table \ref{tab:behaviors Aerostack} shows the set of behaviors of Aerostack grouped in behavior systems. Each robot behavior of this library was programmed to be general in order to be usable for multiple robot applications. The majority of the names used for behaviors follow a particular convention. In general, the first part of the name identifies the \textit{task} performed by the behavior with the base form of the verb (e.g., {\footnotesize\verb|ROTATE|} or {\footnotesize\verb|SELF_LOCALIZE|}). If the same task can be done by different behaviors, the second part of the name (after the word {\footnotesize\verb|WITH|}) distinguishes the \textit{method }used to perform the task (for example, {\footnotesize\verb|ROTATE_WITH_PID_CONTROL|} or {\footnotesize\verb|SELF_LOCALIZE_WITH_EKF_FUSION|}). 

Behaviors are grouped in behavior systems considering two main guidelines. On the one hand, behaviors that share a certain functionality (that can be provided by one or several common ROS packages) are grouped in a system. This happens, for example, with  behaviors about motion that share  MPC control. On the other hand, behaviors are grouped in order to simplify their use (in order to ur a unique ROS launcher for the group of nodes). In this case, the size of the behavior system should not be too large, to avoid launching too many nodes of behaviors that are not used in the final robotic application.

\begin{figure}[htb!]
\begin{center}
\includegraphics[width=0.49\textwidth]{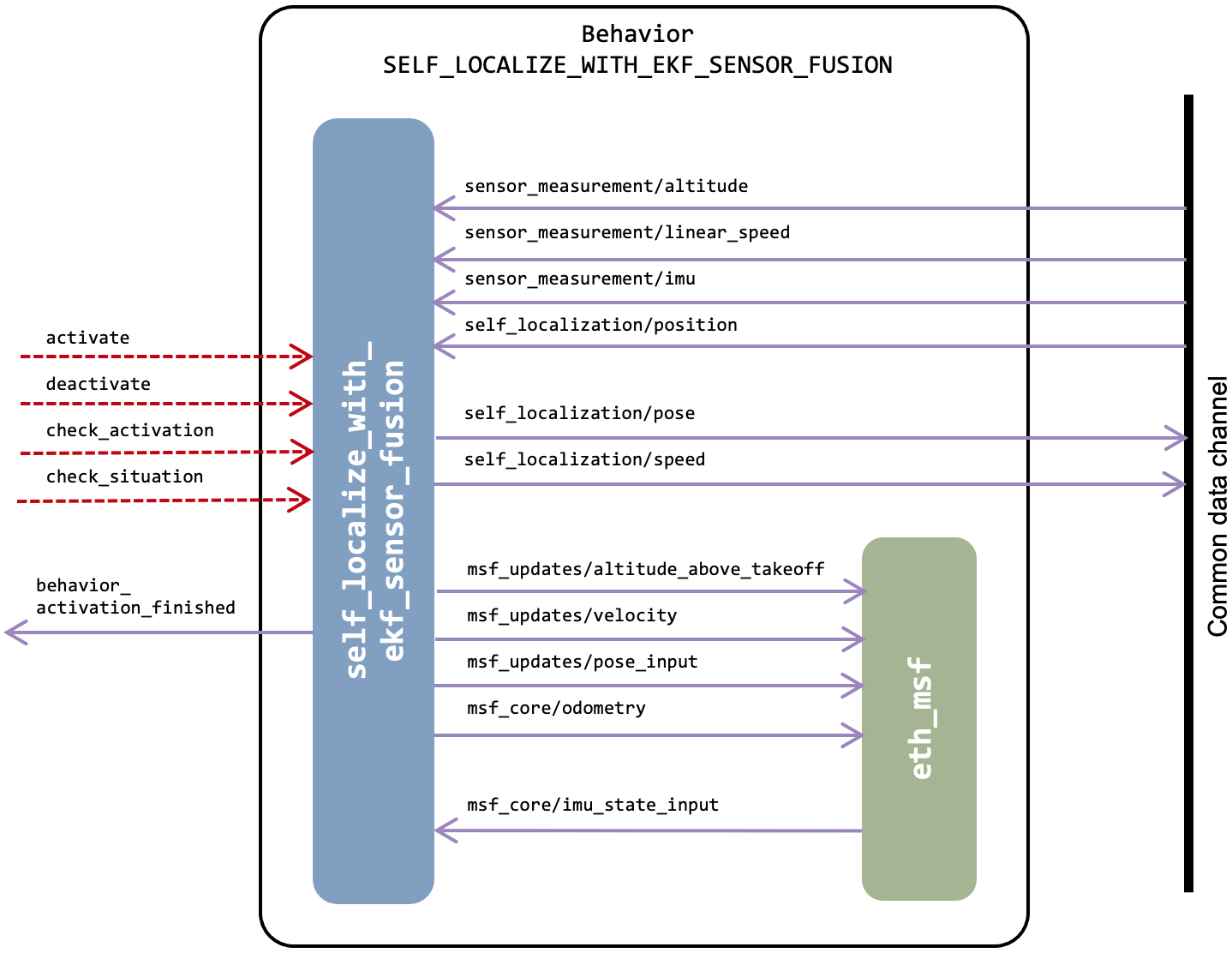}
\caption{ROS-based implementation of the behavior {\scriptsize \textsf{SELF\_LOCALIZE\_ WITH\_SENSOR\_FUSION}}.}
\label{fig:self localize}
\end{center}
\end{figure}

\begin{figure}[htb!]
\begin{center}
\includegraphics[width=0.48\textwidth]{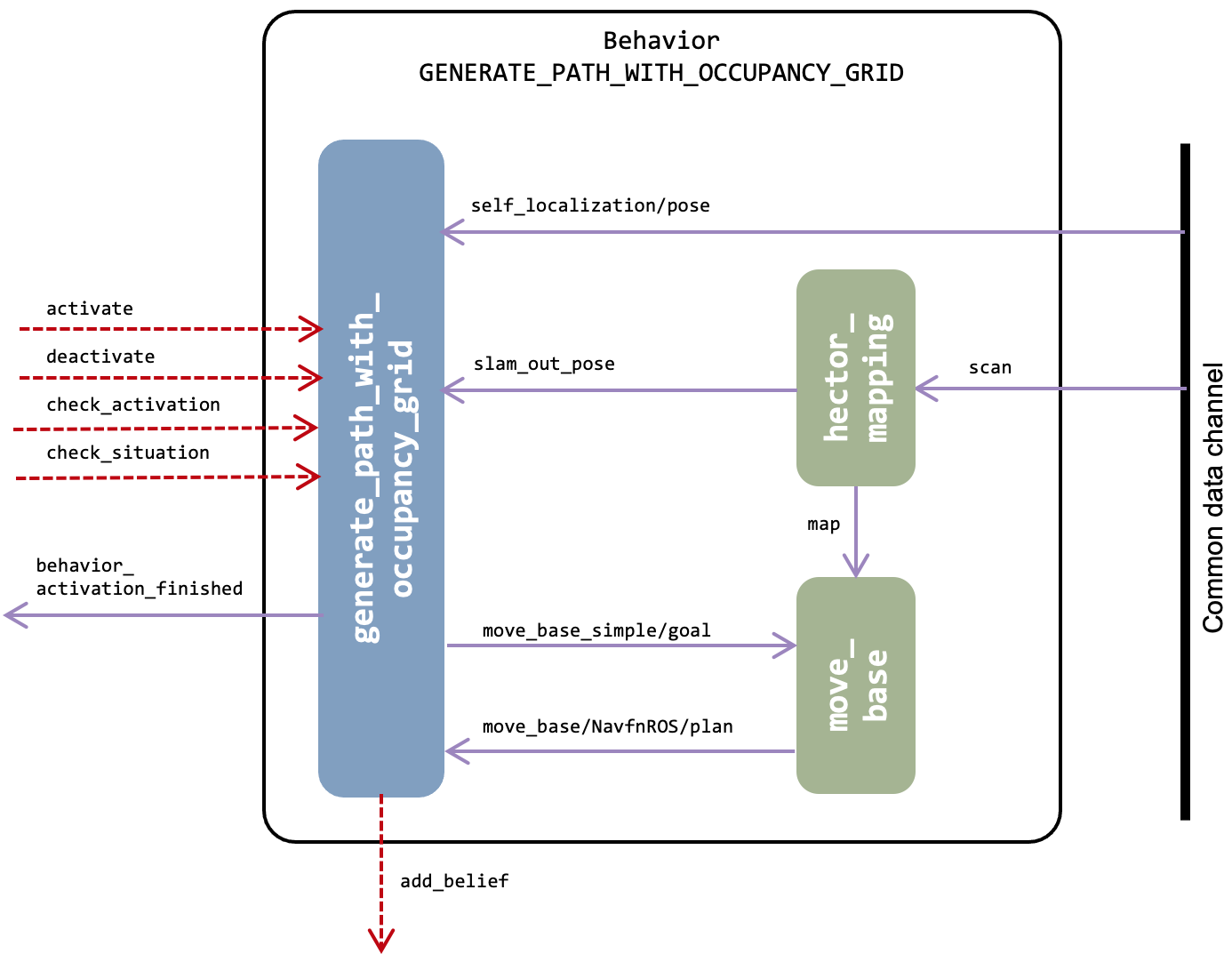}
\caption{ROS-based implementation of the behavior {\scriptsize \textsf{GENERATE\_PATH\_ WITH\_OCCUPANCY\_GRID}}.}
\label{fig:generate path}
\end{center}
\end{figure}

\begin{figure*}[htb!]
\begin{center}
\includegraphics[width=1.0\textwidth]{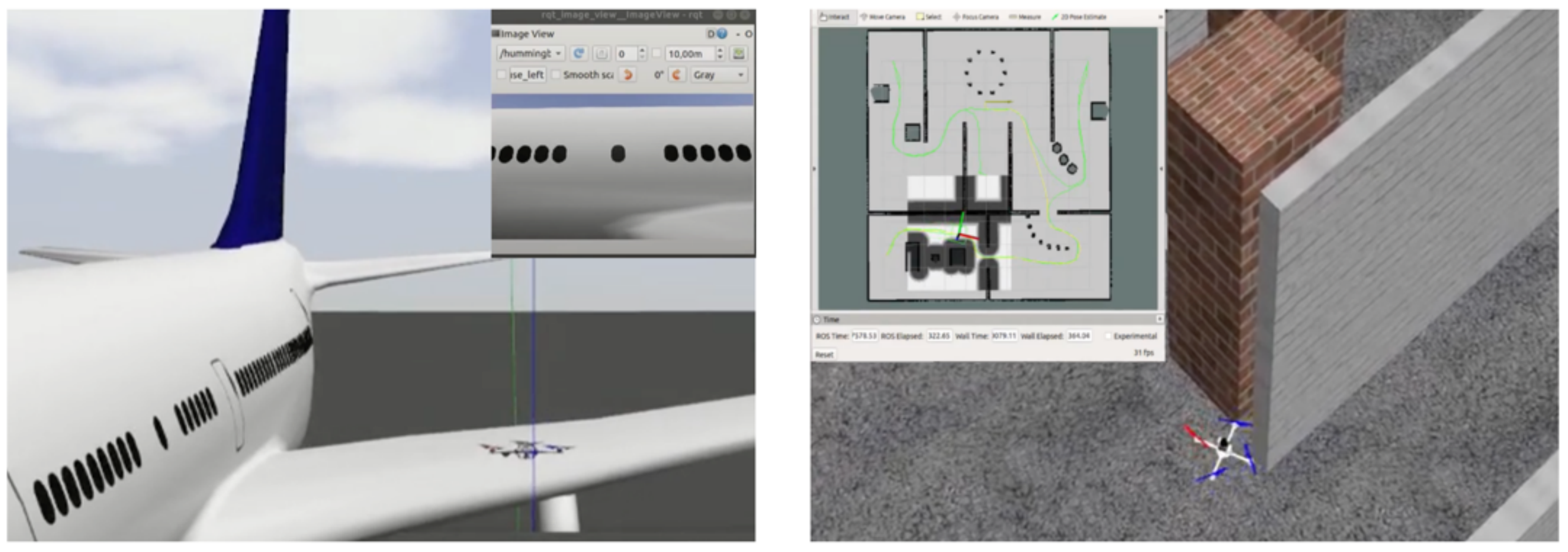}
\caption{Screenshots corresponding to two demonstrative applications of Aerostack 4.0 that were developed with the programming method described in this paper. In one application, an aerial robot moves autonomously around an airplane to inspect its surface with the  help of a camera. In the second application, the aerial robot explores rooms of a building and builds a map. The aerial platform and the environment are simulated using Gazebo.}
\label{fig:application}
\end{center}
\end{figure*}

Figure \ref{fig:self localize} shows the ROS implementation of the behavior called {\footnotesize \verb|SELF_LOCALIZE_WITH_EKF_SENSOR_FUSION|}. The purpose of this behavior is to estimate the localization of the own robot using sensor data and an algorithm based on an Extended Kalman Filter( EKF). This example illustrates an example of recurrent behavior, i.e., a behavior that keeps active its execution until it is deactivated.

This behavior was implemented using the software package \verb|ethzasl_msf|\footnote{\url{https://github.com/ethz-asl/ethzasl_msf}} \cite{Lynen2013} which contains a sensor fusion framework based on an EKF for 6 degrees of freedom pose estimation. Figure \ref{fig:self localize} shows the the particularization of this package for our behavior as a ROS node (in green color) called \verb|eth_msf|. As the figure shows, this node uses specific ROS topics that are not used in the common data channel. Therefore, the behavior execution manager (blue node in the figure) performs the convenient transformations to use the standard topics of the common data channel. 

Figure \ref{fig:generate path} shows the ROS implementation of a behavior called {\footnotesize \verb|GENERATE_PATH_WITH_OCCUPANCY_GRID|}. The goal of this behavior is to generate a path to be followed by the robot to reach a destination. The algorithm for this behavior uses an occupancy grid to represent the presence of obstacles in the environment (detected for example with the help of a lidar sensor) and a path planner to find a path between the current position and the destination. This example illustrates a case of goal-based behavior, i.e., a behavior that finishes its activation when the goal is reached. In this case, the result generated by this behavior (the planned path) is not communicated using the data channel. Instead, this behavior communicates the result using a ROS service that of a node that manages a belief memory (see \cite{Molina2020} to consult details about this belief memory).

The algorithm for this behavior was implemented using the public ROS-based software \textit{ROS Navigation Stack}\footnote{\url{https://github.com/ros-planning/navigation}} \cite{Marder2010} and \textit{Hector Slam}\footnote{\url{https://github.com/tu-darmstadt-ros-pkg/hector_slam}} \cite{Kohlbrecher2011}. Figure \ref{fig:generate path} shows two ROS nodes (in green color) that correspond to these components. 

In this behavior, the execution of the ROS node \verb|move_base| is monitored in the following way. In some situations (e.g., when the computation power is limited), the processing time taken by this node may be longer than expected. The behavior execution manager monitors this processing time and, when it is greater than a timeout threshold, a new request is generated (up to a limited number of requests).

The examples presented above illustrate how behavior execution managers use additional ROS nodes corresponding to public ROS-based software libraries. However, there are other simpler behaviors in Aerostack that are implemented using only a single ROS with all functionalities about execution monitoring, activation management and behavior control. This happens for example with the implementation of simpler behaviors such as {\footnotesize \verb|TAKE OFF|}, {\footnotesize \verb|LAND|} or {\footnotesize \verb|WAIT|}. 

The set of general behaviors presented in this section was validated with the development of demonstrative applications of aerial robotics (13 applications), whose source code was also published with the release 4.0 of the Aerostack software environment.

\subsection{Performance efficiency evaluation}

\begin{table}
\caption{Processing time for different behaviors.}
\label{tab:results performance efficiency} 
\centering
\footnotesize
\begin{tabular}{ | m{3.3cm} | c | c |c |c|} 
\hline
Behavior &  \(a\) & \(t_1\)  & \(t_2\) & \(t_3\)  \\
\hline
\verb|TAKE_OFF| & 1 & 0.6 & 0.2 & 0.6 \\ 
\hline
\verb|LAND| & 1 & 0.4 & 0.1 & 1.1 \\ 
\hline
\verb|SELF_LOCALIZE_WITH_| \verb|GROUND_TRUTH| & 1 & 0.3 & 9.0 & 18.2 \\ 
\hline
\verb|GENERATE_PATH_| \verb|WITH_OCCUPANCY_GRID|& 53 & 3.9 & 7.5 & 3.4 \\ 
\hline
\verb|KEEP_HOVERING_| \verb|WITH_PID_CONTROL|& 41 & 0.3 & 0.8 & 1.4 \\ 
\hline
\verb|FOLLOW_PATH| & 41 & 3.0 & 75.3 & 51.5 \\ 
\hline
\verb|QUADROTOR_PID_| \verb|MOTION_CONTROL| & 1 & 0.1 & 2.6 & 34.2 \\ 
\hline
\verb|QUADROTOR_PID_| \verb|MOTION_CONTROL|  & 1 & 0.1 & 1.6 & 32.8 \\ 
\hline
\end{tabular}
\end{table}

This section presents the results of experiments conducted to evaluate how the programming method presented in this work affects the performance efficiency (e.g., processing time, CPU usage and memory consumption). These experiments were done by using a demonstrative application of Aerostack 4.0 that corresponds to an aerial robot that explores the rooms and builds a map of a building  (see Figure \ref{fig:application}). The characteristics of the computer used for the experiments were the following: CPU AMD Ryzen 7 3800x, 8 cores, 16 threads, 3.9 GHz and 16 GB RAM 3200 MHz.

\begin{table}
\caption{Memory consumption and CPU usage.}
\label{tab:memory consumption} 
\centering
\footnotesize
\begin{tabular}{ | l | c | c | c |} 

\hline
Behavior system & \(b\) & \(MB\) & \(CPU \%\)   \\
\hline
\verb|basic_quadrotor_behaviors| & 7 & 156 & 35 \\ 
\hline
\verb|navigation_with_lidar| & 3 & 139 & 25 \\ 
\hline
\verb|quadrotor_motion_with_pid_control| & 6 & 181 & 50 \\ 
\hline
\end{tabular}
\end{table}

A first set of measures were obtained to analyze the computational overhead produced by the presence of execution monitoring. As it was described, execution monitoring is done by a function that is called periodically with an adjustable frequency. Table \ref{tab:results performance efficiency} shows the results of these tests corresponding to the application for rooms exploration. This application used 8 behaviors. Column \(a\) in the table shows the number of times that each behavior was activated during the mission execution. Some of them were activated only once ({\footnotesize \verb|TAKE_OFF|} and {\footnotesize\verb|LAND|}) and others were activated multiple times (e.g., {\footnotesize\verb|FOLLOW_PATH|} was activated 41 times). Column \(t_1\) in the table shows the average processing time (in microseconds) taken by execution monitoring, which was between 0.1 \(\mu\)s (e.g., for {\footnotesize\verb|LAND|}) and 3.9 \(\mu\)s ({\footnotesize\verb|GENERATE_PATH_WITH_OCCUPANCY_GRID|}). The latter value is higher because it performs more complex monitoring operations (e.g., verifying the execution of the \verb|move_base| node). Column \(t_2\) in the table shows the accumulated processing time (in milliseconds) taken by execution monitoring. This value depends on the number of times execution monitoring was performed. In this mission, execution monitoring was executed at a frequency adjusted to 100 Hz. The maximum accumulated time corresponds to {\footnotesize\verb|FOLLOW_PATH|} with an accumulated value of 75.3 ms. In total, all the time taken by execution monitoring was 97.1 ms (sum of column \(t_2\)). If we compare this time with the time taken to complete the mission (325 seconds) this means the 0.03\% of the total time, which is a low delay generated by execution monitoring.

A second set of measures were obtained to analyze the processing time corresponding to activation management. Column \(t_3\) in Table \ref{tab:results performance efficiency} shows the average time taken (in milliseconds) by activation management for each behavior. The experiments showed that the time taken by this function has a significant variability. For example, the average processing times were the following: 0.6 ms (behavior {\footnotesize\verb|TAKE_OFF|}), 1.1 ms ({\footnotesize\verb|LAND|}), and 51.5 ms ({\footnotesize\verb|FOLLOW_PATH|}). This time depends on each particular application and it is affected by the number of ROS objects created for inter-process communication (e.g., subscribers, publishers and service clients) and the number of processes that are started and stopped. The accumulated time taken by activation management is 2,44 seconds (sum of values of column \(t_3\) multiplied by the values of column \(n\). Considering that the mission was completed in 325 seconds, the delay corresponding to activation management was the 0.75\% of the total time, which is significantly higher than the time taken by execution monitoring. This delay can be admissible in applications that activate behaviors at low average frequencies (less than 0.2 Hz, as it happens in missions used in our experiments). 

Finally, we also estimated the resource utilization (memory consumption and CPU usage) of the ROS nodes corresponding to behavior execution managers. Table \ref{tab:memory consumption} shows these values for different behavior systems. For example, the behavior system \verb|basic_quadrotor_behaviors| has 7 behaviors (column \(b\)), consumes 156 MB of memory and the average CPU usage during the mission is 35\%.  The value of CPU usage corresponds to one of the 16 threads of the computer used for the experiments and the value of memory consumption includes also the memory used by the process used by ROS to launch the execution of groups of nodes (process \textit{roslaunch}).

\FloatBarrier

\section{Conclusion}
In this paper, we have presented a solution for programming behaviors of autonomous robots. This programming approach has been designed to facilitate the coordination of the concurrent execution of multiple behaviors (i.e., basic behaviors that are used for building other more complex robot behaviors).

In the present work, we have identified a set of functions for behavior execution management and we have specified a type of module that encapsulates these functions using uniform interfaces.

One of the contributions of this work is that it uses a decentralized approach for monitoring the execution of each robot behavior. This organization is useful to provide the required flexibility to formulate behavior specific procedures for execution monitoring encapsulated in reusable modules. 

The details of this proposal are formulated in the form of a software library called \verb|behaviorlib| that can be used to program behaviors of different types of robots. This library is implemented in C++ language using inter-processes mechanisms provided by ROS and it is freely available for the community of developers in robotics.

This programming approach has demonstrated its practical utility for building robotic applications. It has been successfully applied to develop programs that implement general aerial robot behaviors. These programs are part of the Aerostack software environment and have been used in the development of multiple aerial robot applications.

\newpage






\ifCLASSOPTIONcaptionsoff
  \newpage
\fi

\IEEEtriggeratref{25}


\bibliographystyle{IEEEtran}
\bibliography{IEEEabrv,references}
%







\end{document}